\documentclass[10pt,twocolumn,letterpaper]{article}

\usepackage{cvpr}
\usepackage{times}
\usepackage{epsfig}
\usepackage{graphicx}
\usepackage{amsmath}
\usepackage{amssymb}

\usepackage[breaklinks=true,bookmarks=false]{hyperref}

\cvprfinalcopy 

\def\cvprPaperID{****} 

\begin{document}

\title{An Empirical Study of Propagation-based Methods for Video Object Segmentation\thanks{Updated on July 30, 2019.}}

\author{Hengkai Guo, Wenji Wang, Guanjun Guo\thanks{equal contribution}, Huaxia Li\footnotemark[2], Jiachen Liu, Qian He, Xuefeng Xiao \\
ByteDance AI Lab \\
}

\maketitle

\begin{abstract}
While propagation-based approaches have achieved state-of-the-art performance for video object segmentation, the literature lacks a fair comparison of different methods using the same settings. In this paper, we carry out an empirical study for propagation-based methods. We view these approaches from a unified perspective and conduct detailed ablation study for core methods, input cues, multi-object combination and training strategies. With careful designs, our improved end-to-end memory networks achieve a global mean of 76.1 on DAVIS 2017 val set.
\end{abstract}

\section{Introduction}
Video object segmentation (VOS) aims at segmenting one or multiple objects for each frame in a video given the annotated masks in the first frame. It is an extremely challenging task due to fast motions, significant appearance changes, severe occlusions and similar dis-tractors.

One important branch for VOS is based on mask propagation \cite{perazzi2017learning} \cite{li2018video}, which estimates the segmentation masks by taking advantage of the masks from previous frame. Recently propagation-based methods have shown continuous performance improvements by introducing more cues such as reference frame information \cite{wug2018fast}, embedding matching \cite{voigtlaender2019feelvos}, or dynamic frame memory \cite{oh2019video}. Among these techniques, different settings in different papers make it difficult to compare them fairly. To our knowledge, there has been little systematically exploration of how different factors and methods influence performance.

In this paper, we analyze several propagation-based methods \cite{perazzi2017learning} \cite{voigtlaender2019feelvos} \cite{oh2019video} in a unified way and conduct extensive empirical experiments to study the impact of different factors and methods on the final performance. We find that input cues, multi-object merging and training paradigms can substantially affect the segmentation in addition to core methods. Based on the findings, we improve the global mean of baseline memory networks\cite{oh2019video} from 74.3 to 76.1 on DAVIS 2017\cite{pont20172017} val set in our implementations. It also achieves a global mean of 67.5 on test-dev set of DAVIS.

\section{Methods}
\begin{figure}[htb]
\centering
{\includegraphics[width=0.48\textwidth]{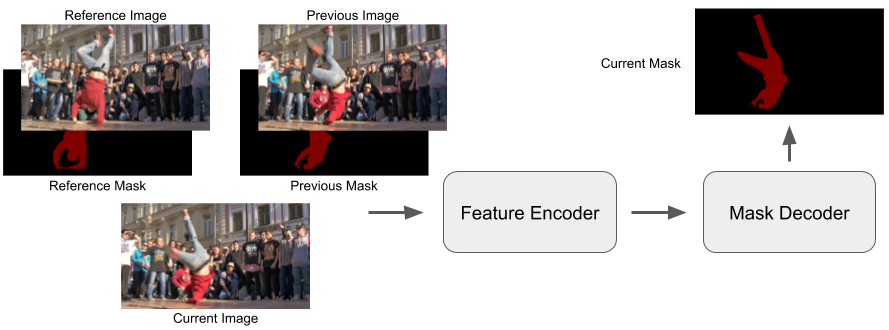}}
\caption{A typical pipeline for a propagation-based system.}
\label{fig_overview}
\end{figure}

A typical propagation-based system \cite{perazzi2017learning} \cite{voigtlaender2019feelvos} \cite{oh2019video} for VOS can be divided into three components: feature encoder, mask decoder and training paradigms. The models first encode inputs from images and masks into features, including feature extraction and fusion. Subsequently masks are predicted by the decoder, often with multi-object combination. To train such models, several strategies such as off-line training and on-line training are adopted. See Fig. \ref{fig_overview} for an overview.

\textbf{Feature encoder.} The feature encoder takes concatenated cues (i.e. images, masks) as inputs and outputs feature maps for mask decoder. Because there may be multiple frame inputs, some methods use correlation matching (FEELVOS\cite{voigtlaender2019feelvos}) or non-local matching (STM\cite{oh2019video}) to combine feature maps from different sources. Correlation matching can be approximately seen as a special case of non-local matching, with the feature maps as keys and the masks as values. We investigate different input cues (previous/reference image and mask) and fusion methods \cite{voigtlaender2019feelvos} \cite{oh2019video}. We also test MaskTrack\cite{perazzi2017learning} with reference and previous cues for a fair comparison. During testing, most of the propagation-based methods\cite{li2018video}\cite{voigtlaender2019feelvos}\cite{oh2019video} use soft probability outputs from previous frames instead of masks. We also examine its effect in our experiments.

\textbf{Mask decoder.} The mask decoder takes the output of feature encoder and predicts the single-object masks independently. Then the masks are merged into a multi-object predictions using object and background probabilities. After merging, the predictions may be normalized and then feed into the next frame. We compare different strategies for computing the background probabilities (constant of 0.5, probability production\cite{li2018video}, or mask tracking \cite{oh2019video}) and normalization methods (sum \cite{li2018video} or logit aggregation \cite{oh2019video}).

\textbf{Training paradigms.} The training procedure contains off-line stage and on-line stage. In off-line stage, the model is trained on the sampled images from videos of VOS datasets. In \cite{wug2018fast}\cite{oh2019video}, static image datasets are used to pre-train the network. Moreover, they train their models using back-propagation through time (BPTT) on videos to obtain training samples that reflect the accumulated errors. For on-line training, first frame annotations of test set are used to fine-tune the off-line model \cite{perazzi2017learning}\cite{li2018video}. BPTT and different on-line fine-tuning patterns (per-dataset\cite{li2018video}\cite{luiten2018premvos} or per-video\cite{xu2019mhp}) are investigated in the experiments.

\textbf{Implementation details.} We employ the DeepLabv3+ architecture \cite{chen2018encoder} (ResNet-50 with output stride of 16) for feature encoder and mask decoder. The concatenated low-level features in decoder are from the query encoder for STM. In all the experiments, we adopt randomly cropped $480\times 480$ patches for training, with batch size of 32. We minimize the cross entropy loss with Adam optimizer for 50 epochs using "poly" policy\cite{chen2018encoder} as learning rate schedule. The learning rate is set to 1e-5 for concatenated inputs, 5e-4 for mask matching, and 5e-5 for non-local feature matching. For fine-tuning, we apply random translation and scaling on the first frame instead of Lucid augmentation\cite{khoreva2018lucid} for simplicity and reproducibility. We generate 2000 samples for each objects in per-dataset fine-tuning and 60000 samples in per-video fine-tuning. And the learning rate is fixed to 1e-5 during on-line training.

\section{Experiments}
\textbf{Datasets.} Our networks are trained on the DAVIS 2017 \cite{pont20172017} training set and YouTube-VOS 2018 \cite{xu2018youtube} training set. In some experiments, only DAVIS dataset is used for faster training. We evaluate our models on DAVIS 2017\cite{pont20172017} validation and test-dev sets, with mean intersection-over-union ($\mathcal{J}$), mean contour accuracy ($\mathcal{F}$) and their mean ($\mathcal{G}$) as in \cite{pont20172017}.

\textbf{Input cues and fusion methods.} We compare different models including MaskTrack\cite{perazzi2017learning}, FEELVOS\cite{voigtlaender2019feelvos} and STM\cite{oh2019video} with different input cues. All the models are trained on two datasets from scratch without fine-tuning. From Table \ref{table_encode}, models with more input cues always achieve better performance because different cues are complementary. Among different fusion methods, non-local matching (STM) significantly outperforms the others on all metrics. This is because STM fuses the information from different frames in space-time dimension, making the encoded features much stronger than other models.

\begin{table}[htb]
\label{table_encode}
\caption{Ablation study of different input cues and fusion methods on DAVIS 2017\cite{pont20172017} validation dataset (\textbf{$\mathcal{M}$}: Last mask. \textbf{$\mathcal{I}$}: Last image. \textbf{Ref}: Reference mask and image.)}
\centering
\begin{tabular}{c|c|c|c|c}
Model & Cues & $\mathcal{G}$ Mean & $\mathcal{J}$ Mean & $\mathcal{F}$ Mean \\
\hline
MSK\cite{perazzi2017learning} & $\mathcal{M}$ & 36.9 & 37.1 & 36.7 \\
& + $\mathcal{I}$ & 46.6 & 43.4 & 49.8 \\
& + Ref & 57.2  & 54.8 & 59.7 \\
\hline
FEEL\cite{voigtlaender2019feelvos} & $\mathcal{M}$ + $\mathcal{I}$ & 46.4 & 44.5 & 48.4 \\
& + Ref & 59.1 & 57.4 & 60.8 \\
\hline
STM\cite{oh2019video} & $\mathcal{M}$ + $\mathcal{I}$ & 64.6 & 62.6 & 66.7 \\
& + Ref & \textbf{74.2} & \textbf{71.7} & \textbf{76.7} \\
\end{tabular}
\end{table}

In Table \ref{table_input_prob} we analyze the effect of the soft input probability during model testing. For FEELVOS and STM, it achieves better global mean with soft input. However, we find that it degrades the results of MaskTrack\cite{perazzi2017learning}. This may because all the mask inputs are in one-hot representation with ground truth for training the models without BPTT. A probability inputs during testing may impair the performance of the model without correlation operator.

\begin{table}[htb]
\label{table_input_prob}
\caption{Ablation study of soft input probability for testing on DAVIS 2017\cite{pont20172017} validation dataset.}
\centering
\begin{tabular}{c|c|c|c|c}
Model & Prob & $\mathcal{G}$ Mean & $\mathcal{J}$ Mean & $\mathcal{F}$ Mean \\
\hline
MSK\cite{perazzi2017learning} & \checkmark & 50.5 & 47.3 & 53.7 \\
 &  & \textbf{52.8} & 48.8 & 56.8 \\
\hline
MSK & \checkmark & 57.2 & 54.8 & 59.7 \\
(+Ref) &  & \textbf{62.0} & 59.5 & 64.5 \\
\hline
FEEL\cite{voigtlaender2019feelvos} & \checkmark & \textbf{59.1} & 57.4 & 60.8 \\
 & & 58.3 & 56.3 & 60.3 \\
\hline
STM\cite{oh2019video} & \checkmark & \textbf{74.2} & 71.7 & 76.7 \\
 & & 73.2 & 70.4 & 76.0 \\
\end{tabular}
\end{table}

\textbf{Multi-object combination.} Table \ref{table_multi_object_test} shows the results of different multi-object combination methods for testing on STM\cite{oh2019video}. Because we do not employ recursive training with BPTT, methods with normalization performs consistently worse than those without it. And the approach to compute background probabilities is not vital for final performance. We simply choose the constant background without normalization for our final models.

\begin{table}[htb]
\label{table_multi_object_test}
\caption{Ablation study of different multi-object combination methods for testing on DAVIS 2017\cite{pont20172017} validation dataset with STM\cite{oh2019video} (\textbf{Const}: constant background probability of 0.5. \textbf{Prod}: product background probability\cite{li2017video}. \textbf{Track}: tracking the background\cite{oh2019video}. \textbf{Sum}: normalized by the sum. \textbf{Logit}: logit aggregation\cite{wug2018fast}).}
\centering
\begin{tabular}{c|c|c|c|c|c}
Bg & Norm & FT & $\mathcal{G}$ Mean & $\mathcal{J}$ Mean & $\mathcal{F}$ Mean \\
\hline
Const & - & \checkmark & 76.1 & 73.5 & 78.8 \\
& & & 74.2 & 71.7 & 76.7 \\
 & Sum & \checkmark & 75.3 & 72.9 & 77.8 \\
& & & 72.8 & 71.0 & 74.6 \\
\hline
Prod & - & \checkmark & \textbf{76.3} & \textbf{73.6} & \textbf{78.9} \\
& & & 74.4 & 71.9 & 76.9 \\
 & Sum\cite{li2018video} & \checkmark & 75.6 & 73.0 & 78.2 \\
& & & 74.2 & 72.1 & 76.3 \\
\hline
Track & - & \checkmark & 76.1 & 73.5 & 78.8 \\
& & & 74.4 & 71.9 & 76.8 \\
 & Sum & \checkmark & 75.3 & 72.7 & 78.0 \\
& & & 74.0 & 71.9 & 76.1 \\
 & Logit\cite{wug2018fast} & \checkmark & 74.8 & 72.0 & 77.5 \\
& & & 74.3 & 72.1 & 76.5 \\
\end{tabular}
\end{table}



\textbf{Training paradigms.} We compare different fine-tuning strategies for different models. For per-video fine-tuning, we only test it with MaskTrack\cite{perazzi2017learning} due to its long testing time. We also provide fine-tuning-only models without training on off-line datasets as baselines. The results are in Table \ref{table_online_train}. Per-dataset fine-tuning can improve the accuracy for MaskTrack \cite{perazzi2017learning} and STM \cite{oh2019video}. The gain from fine-tuning decreases when the base model becomes stronger. And the per-video fine-tuning performs worse than per-dataset fine-tuning in MaskTrack model. This may due to the simple augmentation instead of commonly used Lucid augmentation in other literatures\cite{khoreva2018lucid}\cite{luiten2018premvos}\cite{xu2019mhp}.

\begin{table}[htb]
\label{table_online_train}
\caption{Ablation study of different on-line training paradigms on DAVIS 2017\cite{pont20172017} validation dataset (\textbf{Dataset}: per-dataset fine-tuning. \textbf{Video}: per-video fine-tuning. \textbf{Only}: fine-tuning without mainly training).}
\centering
\begin{tabular}{c|c|c|c|c}
Model & FT & $\mathcal{G}$ Mean & $\mathcal{J}$ Mean & $\mathcal{F}$ Mean \\
\hline
MSK\cite{perazzi2017learning} & - & 50.5 & 47.3 & 53.7 \\
 & Dataset & 66.9 & 64.2 & 69.6 \\
 & Video & 60.8 & 56.9 & 64.8 \\
 & Only & 50.6 & 47.2 & 54.1 \\
\hline
MSK & - & 57.2 & 54.8 & 59.7  \\
(+Ref) & Dataset & 68.2 & 65.7 & 70.7 \\
 & Only & 48.5 & 45.5 & 51.6 \\
\hline
FEEL\cite{voigtlaender2019feelvos} & - & 59.1 & 57.4 & 60.8 \\
 & Dataset & 57.1 & 55.0 & 59.2 \\
 & Only & 33.9 & 32.2 & 35.5 \\
\hline
STM\cite{oh2019video} & - & 74.2 & 71.7 & 76.7 \\
 & Dataset & \textbf{76.1} & \textbf{73.5} & \textbf{78.8} \\
 & Only & 45.4 & 43.2 & 47.7 \\
\end{tabular}
\end{table}

We also experiment with BPTT training (3 frames) and find that it can greatly improve the performance by a large margin in MaskTrack and STM. But due to limited time, we do not apply it on our final models.

\begin{table}[htb]
\label{table_offline_train}
\caption{Ablation study of BPTT on DAVIS 2017\cite{pont20172017} validation dataset with MaskTrack\cite{perazzi2017learning} and STM\cite{oh2019video}.}
\centering
\begin{tabular}{c|c|c|c|c}
Method & BPTT & $\mathcal{G}$ Mean & $\mathcal{J}$ Mean & $\mathcal{F}$ Mean \\
\hline
MSK\cite{perazzi2017learning} & & 35.3 & 35.3 & 35.4 \\
& \checkmark & \textbf{50.5} & \textbf{47.3} & \textbf{53.7} \\
\hline
STM\cite{oh2019video} & & 74.2 & 71.7 & 76.7 \\
& \checkmark & 77.6 & 74.8 & 80.3 \\
\end{tabular}
\end{table}

\textbf{Benchmark.}
According to the conclusions above, we adopt different models with constant multi-object merging and per-dataset fine-tuning. On the DAVIS 2017\cite{pont20172017} validation set (Table \ref{table_benchmark_val}), our STM model beats most of the approaches even without fine-tuning. The only two methods that better than ours are \cite{luiten2018premvos} and \cite{oh2019video}.

\begin{table}[htb]
\label{table_benchmark_val}
\caption{Evaluation on DAVIS 2017\cite{pont20172017} validation set.}
\centering
\begin{tabular}{c|c|c|c|c}
Methods & OL & $\mathcal{G}$ Mean & $\mathcal{J}$ Mean & $\mathcal{F}$ Mean \\
\hline
OSMN\cite{yang2018efficient} & & 54.8 & 52.5 & 57.1 \\
FAVOS\cite{cheng2018fast} & & 58.2 & 54.6 & 61.8 \\
OSVOS\cite{caelles2017one} & \checkmark & 60.3 & 56.6 & 63.9 \\
RGMP\cite{wug2018fast} & & 66.7 & 64.8 & 68.6 \\
CIMN\cite{bao2018cnn} & \checkmark & 67.5 & 64.5 & 70.5 \\
OnAVOS\cite{voigtlaender2017online} & \checkmark & 67.9 & 64.5 & 71.2 \\
OSVOS-S\cite{maninis2018video} & \checkmark & 68.0 & 64.7 & 71.3 \\
AGAME\cite{johnander2018generative} & & 70.0 & 67.2 & 72.7 \\
FEELVOS\cite{voigtlaender2019feelvos} & & 71.5 & 69.1 & 74.0 \\
DOL\cite{robinson2019discriminative} & \checkmark & 73.4 & 71.3 & 75.5 \\
DyeNet\cite{li2018video} & \checkmark & 74.1 & - & - \\
MHP-VOS\cite{xu2019mhp} & \checkmark & 75.3 & 71.8 & 78.8 \\
PReMVOS\cite{luiten2018premvos} & \checkmark & 77.8 & 73.9 & 81.7 \\
STM\cite{oh2019video} & & \textbf{81.8} & \textbf{79.2} & \textbf{84.3} \\
\hline
Ours (MSK Ref) & & 57.2 & 54.8 & 59.7 \\
Ours (MSK Ref + FT) & \checkmark & 68.2 & 65.7 & 70.7 \\
Ours (FEEL) & & 59.1 & 57.4 & 60.8 \\
Ours (FEEL + FT) & \checkmark & 57.1 & 55.0 & 59.2 \\
Ours (STM) &  & 74.2 & 71.7 & 76.7 \\
Ours (STM + FT) & \checkmark & \textbf{76.1} & \textbf{73.5} & \textbf{78.8} \\
\end{tabular}
\end{table}

As for test-dev set (Table \ref{table_benchmark_testdev}), our STM model without fine-tuning also performs better than other off-line models. However, it is not able to reach the other results with on-line fine-tuning\cite{li2018video}\cite{xu2019mhp}\cite{luiten2018premvos}.

\begin{table}[htb]
\label{table_benchmark_testdev}
\caption{Evaluation on DAVIS 2017\cite{pont20172017} test-dev set.}
\centering
\begin{tabular}{c|c|c|c|c}
Methods & OL & $\mathcal{G}$ Mean & $\mathcal{J}$ Mean & $\mathcal{F}$ Mean \\
\hline
OSMN\cite{yang2018efficient} & & 41.3 & 37.7 & 44.9 \\
FAVOS\cite{cheng2018fast} & & 43.6 & 42.9 & 44.2 \\
OSVOS\cite{caelles2017one} & \checkmark & 50.9 & 47.0 & 54.8 \\
RGMP\cite{wug2018fast} & & 52.9 & 51.4 & 54.4 \\
OnAVOS\cite{voigtlaender2017online} & \checkmark & 56.5 & 53.4 & 59.6 \\
OSVOS-S\cite{maninis2018video} & \checkmark & 57.5 & 52.9 & 62.1 \\
FEELVOS\cite{voigtlaender2019feelvos} & & 57.8 & 55.2 & 60.5 \\
DyeNet\cite{li2018video} &  & 62.5 & 60.2 & 64.8 \\
VS-ReID\cite{li2017video} & \checkmark & 66.1 & 64.4 & 67.8 \\
CIMN\cite{bao2018cnn} & \checkmark & 67.5 & 64.5 & 70.5 \\
DyeNet\cite{li2018video} & \checkmark  & 68.2 & 65.8 & 70.5 \\
MHP-VOS\cite{xu2019mhp} & \checkmark & 69.5 & 66.4 & 72.7 \\
PReMVOS\cite{luiten2018premvos} & \checkmark & \textbf{71.6} & \textbf{67.5} & \textbf{75.7} \\
\hline
Ours (STM) &  & 63.3 & 60.5 & 66.1 \\
Ours (STM + FT) & \checkmark & 67.5 & 64.3 & 70.7 \\
\end{tabular}
\end{table}

For the 2019 DAVIS Challenge, our submission achieved a global mean of 69.2.

\section{Conclusions}
In conclusion, we present an empirical study of propagation-based methods for VOS to analyze the influence of different factors and models. Our results suggest that memory network\cite{oh2019video} with more input cues, constant background probability without normalization and per-dataset fine-tuning can achieve better performance for this task.

{\small
\bibliographystyle{ieee_fullname}
\bibliography{egbib}

\begin{thebibliography}{10}\itemsep=-1pt

\bibitem{bao2018cnn}
Linchao Bao, Baoyuan Wu, and Wei Liu.
\newblock Cnn in mrf: Video object segmentation via inference in a cnn-based
  higher-order spatio-temporal mrf.
\newblock In {\em Proceedings of the IEEE Conference on Computer Vision and
  Pattern Recognition}, pages 5977--5986, 2018.

\bibitem{caelles2017one}
Sergi Caelles, Kevis-Kokitsi Maninis, Jordi Pont-Tuset, Laura Leal-Taix{\'e},
  Daniel Cremers, and Luc Van~Gool.
\newblock One-shot video object segmentation.
\newblock In {\em Proceedings of the IEEE conference on computer vision and
  pattern recognition}, pages 221--230, 2017.

\bibitem{chen2018encoder}
Liang-Chieh Chen, Yukun Zhu, George Papandreou, Florian Schroff, and Hartwig
  Adam.
\newblock Encoder-decoder with atrous separable convolution for semantic image
  segmentation.
\newblock In {\em Proceedings of the European Conference on Computer Vision
  (ECCV)}, pages 801--818, 2018.

\bibitem{cheng2018fast}
Jingchun Cheng, Yi-Hsuan Tsai, Wei-Chih Hung, Shengjin Wang, and Ming-Hsuan
  Yang.
\newblock Fast and accurate online video object segmentation via tracking
  parts.
\newblock In {\em Proceedings of the IEEE Conference on Computer Vision and
  Pattern Recognition}, pages 7415--7424, 2018.

\bibitem{johnander2018generative}
Joakim Johnander, Martin Danelljan, Emil Brissman, Fahad~Shahbaz Khan, and
  Michael Felsberg.
\newblock A generative appearance model for end-to-end video object
  segmentation.
\newblock {\em arXiv preprint arXiv:1811.11611}, 2018.

\bibitem{khoreva2018lucid}
Anna Khoreva, Rodrigo Benenson, Eddy Ilg, Thomas Brox, and Bernt Schiele.
\newblock Lucid data dreaming for video object segmentation.
\newblock {\em International Journal of Computer Vision}, pages 1--23, 2018.

\bibitem{li2018video}
Xiaoxiao Li and Chen Change~Loy.
\newblock Video object segmentation with joint re-identification and
  attention-aware mask propagation.
\newblock In {\em Proceedings of the European Conference on Computer Vision
  (ECCV)}, pages 90--105, 2018.

\bibitem{li2017video}
Xiaoxiao Li, Yuankai Qi, Zhe Wang, Kai Chen, Ziwei Liu, Jianping Shi, Ping Luo,
  Xiaoou Tang, and Chen~Change Loy.
\newblock Video object segmentation with re-identification.
\newblock {\em arXiv preprint arXiv:1708.00197}, 2017.

\bibitem{luiten2018premvos}
Jonathon Luiten, Paul Voigtlaender, and Bastian Leibe.
\newblock Premvos: Proposal-generation, refinement and merging for video object
  segmentation.
\newblock {\em arXiv preprint arXiv:1807.09190}, 2018.

\bibitem{maninis2018video}
K-K Maninis, Sergi Caelles, Yuhua Chen, Jordi Pont-Tuset, Laura Leal-Taix{\'e},
  Daniel Cremers, and Luc Van~Gool.
\newblock Video object segmentation without temporal information.
\newblock {\em IEEE transactions on pattern analysis and machine intelligence},
  41(6):1515--1530, 2018.

\bibitem{oh2019video}
Seoung~Wug Oh, Joon-Young Lee, Ning Xu, and Seon~Joo Kim.
\newblock Video object segmentation using space-time memory networks.
\newblock {\em arXiv preprint arXiv:1904.00607}, 2019.

\bibitem{perazzi2017learning}
Federico Perazzi, Anna Khoreva, Rodrigo Benenson, Bernt Schiele, and Alexander
  Sorkine-Hornung.
\newblock Learning video object segmentation from static images.
\newblock In {\em Proceedings of the IEEE Conference on Computer Vision and
  Pattern Recognition}, pages 2663--2672, 2017.

\bibitem{pont20172017}
Jordi Pont-Tuset, Federico Perazzi, Sergi Caelles, Pablo Arbel{\'a}ez, Alex
  Sorkine-Hornung, and Luc Van~Gool.
\newblock The 2017 davis challenge on video object segmentation.
\newblock {\em arXiv preprint arXiv:1704.00675}, 2017.

\bibitem{robinson2019discriminative}
Andreas Robinson, Felix~J{\"a}remo Lawin, Martin Danelljan, Fahad~Shahbaz Khan,
  and Michael Felsberg.
\newblock Discriminative online learning for fast video object segmentation.
\newblock {\em arXiv preprint arXiv:1904.08630}, 2019.

\bibitem{voigtlaender2019feelvos}
Paul Voigtlaender, Yuning Chai, Florian Schroff, Hartwig Adam, Bastian Leibe,
  and Liang-Chieh Chen.
\newblock Feelvos: Fast end-to-end embedding learning for video object
  segmentation.
\newblock {\em arXiv preprint arXiv:1902.09513}, 2019.

\bibitem{voigtlaender2017online}
Paul Voigtlaender and Bastian Leibe.
\newblock Online adaptation of convolutional neural networks for the 2017 davis
  challenge on video object segmentation.
\newblock In {\em The 2017 DAVIS Challenge on Video Object Segmentation-CVPR
  Workshops}, volume~5, 2017.

\bibitem{wug2018fast}
Seoung Wug~Oh, Joon-Young Lee, Kalyan Sunkavalli, and Seon Joo~Kim.
\newblock Fast video object segmentation by reference-guided mask propagation.
\newblock In {\em Proceedings of the IEEE Conference on Computer Vision and
  Pattern Recognition}, pages 7376--7385, 2018.

\bibitem{xu2018youtube}
Ning Xu, Linjie Yang, Yuchen Fan, Dingcheng Yue, Yuchen Liang, Jianchao Yang,
  and Thomas Huang.
\newblock Youtube-vos: A large-scale video object segmentation benchmark.
\newblock {\em arXiv preprint arXiv:1809.03327}, 2018.

\bibitem{xu2019mhp}
Shuangjie Xu, Daizong Liu, Linchao Bao, Wei Liu, and Pan Zhou.
\newblock Mhp-vos: Multiple hypotheses propagation for video object
  segmentation.
\newblock {\em arXiv preprint arXiv:1904.08141}, 2019.

\bibitem{yang2018efficient}
Linjie Yang, Yanran Wang, Xuehan Xiong, Jianchao Yang, and Aggelos~K
  Katsaggelos.
\newblock Efficient video object segmentation via network modulation.
\newblock In {\em Proceedings of the IEEE Conference on Computer Vision and
  Pattern Recognition}, pages 6499--6507, 2018.

\end{thebibliography}
}

\end{document}